\PassOptionsToPackage{dvipsnames}{xcolor}
\documentclass{article}
\usepackage{arxiv}
\usepackage{natbib}[numbers,sort,compress]

\def\eqref#1{(\ref{#1})}

\usepackage{wrapfig}

\usepackage{multirow,mathtools }

\usepackage{pifont}
\usepackage{color, colortbl}

\usepackage{blindtext}
\usepackage{lipsum}

\usepackage{multirow}
\usepackage{listings}

\usepackage{bbm}

\usepackage [english]{babel}
\usepackage[autostyle, english = american]{csquotes}

\usepackage{pifont}
\usepackage{url}
\usepackage[most]{tcolorbox}

\usepackage{lipsum}
\usepackage{soul}
\usepackage{xcolor}
\usepackage{wrapfig}
\usepackage{multirow,mathtools } 

\usepackage{adjustbox}
\MakeOuterQuote{"}

\usepackage{microtype}
\usepackage{graphicx}
\usepackage{subfigure}
\usepackage{booktabs}

\usepackage{hyperref}

\usepackage{amsmath}

\usepackage[capitalize,noabbrev]{cleveref}

\usepackage{nicefrac}

\usepackage{tablefootnote}

\usepackage{float} 
\usepackage{diagbox}

\usepackage{pifont}
\newcommand{\cmark}{\textcolor{green}{\ding{51}}}

\usepackage{color, colortbl}
\definecolor{Gray}{gray}{0.93}
\definecolor{Orange}{rgb}{1,0.5,0}
\definecolor{DGray}{gray}{0.83}
\definecolor{LightCyan}{rgb}{0.88,1,1}

\definecolor{WarnREd}{rgb}{1,0.4,0.4}
\definecolor{WarnOrange}{rgb}{1,0.682,0.502}
\definecolor{WarnPink}{rgb}{0.9176, 0.7215, 0.7215}
\definecolor{GoodGreen}{rgb}{0.5019, 0.9215, 0.6039}

\usepackage[T1]{fontenc}

\definecolor{styleblue}{HTML}{504099}
\definecolor{mypurple}{HTML}{9391ff}

\definecolor{bluegray}{rgb}{0.4, 0.6, 0.8}
\definecolor{ceruleanblue}{rgb}{0.16, 0.32, 0.75}

\hypersetup{
colorlinks=true,
citecolor=ceruleanblue,
linkcolor=ceruleanblue,
urlcolor=black
}

\definecolor{LightCyan}{rgb}{0.88,1,1}

\usepackage{pdflscape}
\usepackage{adjustbox}
\usepackage{makecell}
\usepackage{multirow}
\usepackage{colortbl}
\usepackage[table]{xcolor}

\usepackage{amsmath,amsfonts,bm}

\def\eqref#1{(\ref{#1})}

\def\1{\bm{1}}

\DeclareMathAlphabet{\mathsfit}{\encodingdefault}{\sfdefault}{m}{sl}
\SetMathAlphabet{\mathsfit}{bold}{\encodingdefault}{\sfdefault}{bx}{n}

\DeclareMathOperator*{\minimize}{\text{minimize}}

\newcommand{\btheta}{{\boldsymbol{\theta}}}

\definecolor{BS_color}{RGB}{91,59,140}

\definecolor{Sijia_color}{rgb}{0.858, 0.188, 0.478}

\usepackage{times}
\usepackage{latexsym}

\usepackage[T1]{fontenc}

\usepackage{microtype}

\usepackage{inconsolata}
\usepackage{fontawesome}

\usepackage{graphicx}
\usepackage{booktabs}
\usepackage{float} 
\usepackage{adjustbox}
\usepackage{multirow}
\usepackage{wrapfig} 

\everydisplay{\small}

\title{Can Native Safety Travel? Reusing Safety Pretraining Across Models  via On-Policy Distillation}

\author{%
Bingqi Shang$^{1}$\quad Yang Zhang$^{2}$\quad Yiwei Chen$^{1}$\quad Changsheng Wang$^{1}$\\
\textbf{Shiyu Chang$^{3}$\quad Bingquan Shen$^{4}$\quad Sijia Liu$^{1,2}$}\\
$^{1}$Michigan State University \quad
$^{2}$MIT-IBM Watson AI Lab\\
$^{3}$University of California, Santa Barbara \quad
$^{4}$National University of Singapore
\\[0.6em]
\href{https://github.com/OPTML-Group/Safety-Pretrain-Distillation}{\faGithub\ GitHub}
}
\date{}

\begin{document}
\maketitle

\begin{abstract}
Safety pretraining has emerged as a promising paradigm for embedding safety \textit{deeply} into a model during pretraining, yielding robustness to subsequent fine-tuning  that conventional safety post-training often lacks.  We refer to this safety acquired during model pretraining as \emph{native safety}. Yet native safety is expensive to acquire, confined to the model in which it was pretrained, and often accompanied by lower general utility. In this work, we rethink the practical value of safety-pretrained models and introduce, for the first time, the \emph{native safety reuse problem}: Can native safety be transferred from a specialized, potentially smaller, safety-pretrained ``teacher'' to standard-pretrained ``students,'' enabling them to inherit its robustness while preserving their general utility? We show that native safety can ``travel'' through on-policy distillation. A standard-pretrained student generates responses to safety-pretraining prompts, while a small safety-pretrained teacher supervises the student's token-level predictions along its own generated trajectories; cross-tokenizer alignment further enables transfer across model families. The resulting students exhibit substantially greater safety robustness than their safety-post-trained counterparts while retaining higher general utility than the teacher: subsequent fine-tuning raises the average attack success rate (ASR) of the 1.7B student by only 5.00\%, compared with 11.58\% for safety post-training, and its average utility of 48.20\% far exceeds the aligned teacher's 27.42\%.
Unlike conventional large-to-small knowledge distillation, our framework enables selective, small-to-large safety transfer: it reuses the robust safety expertise of a smaller pretrained teacher without inheriting its utility limitations. 
It further enables native safety to ``travel'' to broader applications, including robust machine unlearning and vision-language model safety alignment.
These findings establish native safety as a reusable model property rather than one permanently tied to the originally pretrained model.
\end{abstract}

\section{Introduction}
\label{sec:intro}

Large language models (LLMs) are increasingly deployed in open-ended and high-stakes settings, making their safety a priority. The prevailing approach is \emph{safety alignment through post-training}, which uses supervised fine-tuning, preference optimization, reinforcement learning, or related techniques to shape the behavior of a pretrained model \citep{ouyang2022training,rafailov2023direct,bianchi2024safety,zou2024improving,liu2025rethinking,jia2025beyond,ji2023ai,shen2023large}.
Although safety post-training can substantially reduce harmful compliance at the resulting checkpoint, this protection is often fragile. Aligned models remain vulnerable to jailbreaking attacks \citep{zou2023universal,wei2023jailbroken,chao2025jailbreaking,qi2025safety}.
More alarmingly, their safety can deteriorate during subsequent fine-tuning even when the downstream data are entirely \textit{benign} and unrelated to harmful behavior  \citep{qi2024fine,he2024your,huang2024harmful,wang2025invariance}.
Thus, ordinary model adaptation can inadvertently undo safety previously acquired through post-training.
 
From the perspective of downstream adaptation, the difficulty of preserving post-training safety has been viewed as a fine-tuning-induced misalignment problem \citep{siddiqui2026position}. However, it also exposes a fundamental limitation of safety post-training itself: the learned safety may remain \emph{shallow} \citep{qi2025safety,qi2024fine,tice2026alignment}. Post-training updates an already pretrained model using a comparatively small alignment dataset and limited optimization budget, often inducing only modest changes to its weights \citep{siddiqui2026dormant,liu2019towards}. Consequently, the model may learn safe behavior at the surface level without deeply integrating safety into its underlying representations \citep{qi2025safety}. Subsequent safety-irrelevant updates can then overwrite this behavior when improving performance on the intended downstream task, such as mathematical reasoning. 

\begin{wrapfigure}[26]{r}{0.7\textwidth}
\vspace*{-7mm}
\centering
  \includegraphics[width=\linewidth]{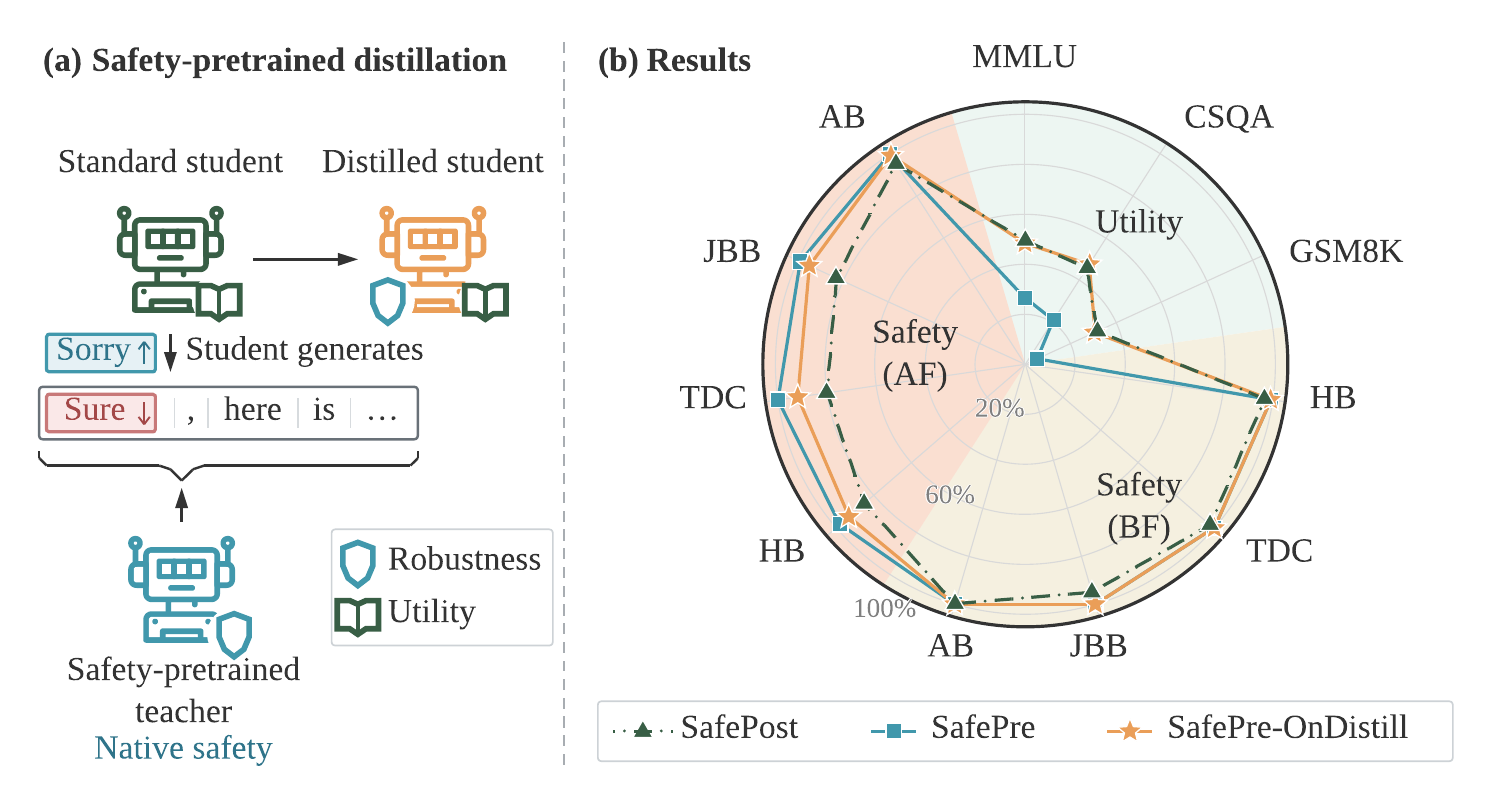}
  \vspace*{-10mm}
  \caption{\small{
  Overview of proposed native safety reuse. \textbf{(a)} On-policy safety distillation: 
  The standard-pretrained student generates responses to safety prompts, while the safety-pretrained teacher supervises its token-level predictions along the generated trajectories, enabling the distilled student to inherit native safety while preserving its general utility.
  \textbf{(b)} Performance overview: The distilled student (from the standard model \texttt{SmolLM2-1.7B}) is more robust to subsequent fine-tuning  on GSM8K than its safety-post-trained counterpart while retaining substantially higher utility than the safety-pretrained teacher (\texttt{SafeLM-1.7B}). Utility is measured by accuracy on MMLU, CommonsenseQA (CSQA), and GSM8K; Safety is measured by the complement of the attack success rate (ASR), \textit{i.e.}, $(100-\mathrm{ASR})\%$, on HarmBench (HB), TDC, JailbreakBench (JBB), and AdvBench (AB), both before (BF) and after (AF) fine-tuning. Higher values are better on all axes.
  For a fair comparison, all models, teacher included, are evaluated after the same safety post-training for instruction following and safety alignment.
  } 
  }  
    \vspace*{-3mm}
  \label{fig:overview}  
  \end{wrapfigure}
To overcome the limitations of safety post-training, \emph{safety pretraining} has recently emerged as a promising paradigm for embedding safety more deeply into a model \citep{korbak2023pretraining,maini2026safety,sam2026should,o2026deep,roland2026modular,stranisci2026they,tice2026alignment}.
This paradigm integrates safety-relevant data and learning objectives directly into the pretraining process. Recent studies show that safety-pretrained models preserve their safety substantially better under subsequent fine-tuning than models aligned solely through post-training. We refer to this deeply embedded and adaptation-resilient property as \textbf{native safety}: safety acquired during pretraining, characterized not only by safe behavior at the current checkpoint, but also by robustness to subsequent fine-tuning.

\vspace*{3mm}

Despite its promise, safety pretraining faces some practical limitations. First, pretraining a model from scratch is substantially computationally expensive, making it impractical to repeat for every architecture, scale, or application. Second, native safety is confined to the particular model in which it was pretrained and cannot directly benefit the broader ecosystem of open-weight models.
Third, stronger native safety may come at the cost of lower general utility \citep{maini2026safety}, and thus, directly deploying a safety-pretrained model can  sacrifice capabilities already available in standard models. These limitations motivate the central question of this work:
\begin{tcolorbox}[before skip=2mm, after skip=0.0cm, boxsep=0.0cm, middle=0.0cm, top=0.05cm, bottom=0.05cm, boxrule=0.6pt]
\begin{center}
\textbf{(Q)} \textit{Can native safety ``travel''---that is, be transferred from a safety-pretrained model and reused by larger, more capable standard models?}  
     \end{center}
\end{tcolorbox} 
\vspace*{2mm}

We refer to \textbf{(Q)} as the \textbf{native safety reuse problem}. Solving it can improve the value and scalability of safety pretraining: rather than treating native safety as a model-specific property that must be relearned through costly pretraining, we could acquire it once and reuse it across architectures and modalities. Our objective differs fundamentally from conventional knowledge distillation~\citep{hinton2015distilling,gou2021knowledge,kim2016sequence,hsieh2023distilling,gu2023minillm,xu2024survey,song2026survey}, which typically transfers the general capabilities of a large teacher to a smaller student for efficient deployment.
Instead, a particularly valuable setting in our work is \emph{small-to-large transfer}, where a smaller safety-pretrained teacher transfers its native safety expertise to a larger, more capable student without compromising the student's general utility.

Our central finding is that native safety travels most effectively \emph{on-policy}: the student generates responses to safety-pretraining prompts, and the teacher supervises its token-level predictions along these trajectories, directly correcting the states and failure modes encountered by the student (see \textbf{Fig.\,\ref{fig:overview}}  for an overview). To enable reuse across model families, we further address tokenizer mismatch through cross-tokenizer on-policy distillation~\citep{patiño2025_unlocking_on_policy_distillation_for_any_model_family,niu2026breaking,sun2026simct}, which aligns token sequences and vocabulary distributions before transferring teacher supervision. In our study, this allows the small \texttt{SafeLM-1.7B} teacher \citep{maini2026safety} to improve larger students from another model family, including \texttt{Llama-3.2-3B} and \texttt{Llama-3-8B}~\citep{grattafiori2024llama}.
Beyond LLM safety alignment, we demonstrate native safety reuse in machine unlearning, where it makes unlearning more robust, and in multimodal safety alignment, where a language-based safety-pretrained teacher enhances the safety robustness of a vision-language model (VLM) without modifying its visual components. Our \textbf{main contributions} are summarized as follows:

 $\bullet$ We introduce the \emph{native safety reuse problem}, reframing safety pretraining as a reusable resource for improving other models rather than a costly, model-bound outcome.

 $\bullet$ We show that native safety travels effectively through on-policy distillation, requiring both a safety-pretrained teacher and on-policy supervision. Cross-tokenizer alignment further enables small-to-large transfer across model scales and architectures, allowing \texttt{SafeLM-1.7B} to improve larger \texttt{Llama} students without inheriting the teacher's utility limitations.

 $\bullet$ Across multiple safety and   general-utility benchmarks, distilled students exhibit greater robustness to subsequent fine-tuning than their safety-post-trained counterparts while retaining higher utility than the safety-pretrained teacher.
 See Fig.\,\ref{fig:overview}(b) for some experimental highlights.
 
 $\bullet$ We extend native safety reuse to robust machine unlearning and VLM safety alignment, demonstrating transfer beyond the teacher's original architecture and task.

\section{Related Work}
\noindent
\textbf{Safety pretraining and robustness to subsequent fine-tuning.}
Safety pretraining embeds safety into a model by rewriting or filtering its pretraining corpus~\citep{maini2026safety,o2026deep,stranisci2026they}, controlling the timing and structure of safety data~\citep{sam2026should,tice2026alignment,roland2026modular}, or conditioning training on preference scores~\citep{korbak2023pretraining}. \citet{maini2026safety} show that the resulting safety-pretrained model, \texttt{SafeLM}, preserves safety under subsequent fine-tuning, whereas safety introduced through post-training can erode even on benign data~\citep{qi2024fine,he2024your,huang2024harmful}. A complementary line of work protects post-trained models during fine-tuning~\citep{huang2024vaccine,ham2025safety} or makes their weights resistant to tampering~\citep{siddiqui2026dormant,wang2025invariance}. These approaches protect only the model being pretrained or defended. In contrast, we ask whether robustness acquired during safety pretraining can be transferred or reused by a standard-pretrained student, and improve the safety robustness of the resulting student under subsequent fine-tuning without additional defenses.

\noindent
\textbf{On-policy and cross-tokenizer distillation.}
On-policy distillation trains a student on its own rollouts under teacher supervision~\citep{lin2020autoregressive,agarwal2024policy,gu2023minillm,song2026survey}. Its effectiveness can depend more on teacher--student proximity than on teacher strength~\citep{li2026rethinking,zhu2026many}, and can improve by matching the teacher's top-$k$ distribution rather than only sampled tokens~\citep{fu2026revisiting}. Cross-tokenizer distillation methods align predictive distributions across vocabularies through optimal transport~\citep{boizard2024towards}, sequence and vocabulary alignment~\citep{patiño2025_unlocking_on_policy_distillation_for_any_model_family}, or finer-grained aligned units~\citep{niu2026breaking,sun2026simct}. Prior work primarily studies capability transfer, including weak-to-strong settings in which smaller models supervise larger ones~\citep{burns2023weak,yu2026weak}. In contrast, we distill safety from a small safety-pretrained teacher to students of equal or larger scale, and study whether the transferred safety survives subsequent fine-tuning.

\noindent
\textbf{Safety distillation.}
Safety post-training can distill refusal behavior from aligned LLMs~\citep{bianchi2024safety,jiang2024wildteaming}, although off-policy distillation from an aligned teacher may yield a less safe student~\citep{zhang2025response,jahan2025black}. On-policy distillation has also recently been applied to safety alignment: supervising the student's own rollouts lowers the safety tax~\citep{fu2026reducing}, and restricting the supervision to safety-critical tokens further limits the utility loss~\citep{li2026safesteer}. In these methods the teacher is the student itself under a safety prompt, a constitution, or a steering vector~\citep{fu2026reducing,wen2026constitutional,li2026safesteer}, so the distilled safety is bounded by that of the prompted student~\citep{zhu2026many}. Other approaches employ a smaller corrector to align a larger model~\citep{ji2024aligner} or use distillation to make unlearning resistant to relearning~\citep{lee2026distillation}. Yet,  little work has studied distillation from safety-pretrained models.

\section{Problem and Motivation: Native Safety through Pretraining}
\label{sec:formulation}

\noindent\textbf{Safety pretraining: A costly yet more robust alternative to safety post-training.}
As a concrete setup, we consider the safety-pretraining framework of \citet{maini2026safety} and adopt their released \textit{safety-pretrained model} \texttt{SafeLM-1.7B} in our subsequent study (whose \textit{standard-pretrained counterpart} is \texttt{SmolLM2-1.7B}).
Their pipeline constructs a rich safety-pretraining corpus from web documents in the SmolLM2 pretraining corpus \citep{allal2025smollm2}, which are then scored according to their safety risk. Moderately unsafe documents are rephrased as safe, context-rich text (\textsc{SafeWeb}), whereas highly unsafe documents are transformed into refusal dialogues (\textsc{RefuseWeb}) and rewritten as educational articles (\textsc{MoralEducation}).  
By contrast, the standard \textit{safety post-training} pipeline considered by \citet{maini2026safety} relies on a comparatively limited collection of instruction-following and alignment datasets, including   \textsc{UltraChat-200k}  \citep{ding2023enhancing}, \textsc{WildGuardMix} \citep{han2024wildguard}, and \textsc{WildJailbreak} \citep{jiang2024wildteaming}.

\begin{wrapfigure}{r}{0.45\textwidth}
  \centering
  \vspace*{-3mm}
\includegraphics[width=\linewidth]{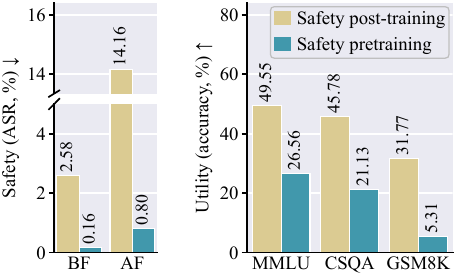} 
\vspace*{-7mm}
  \caption{\small{Safety pretraining largely preserves safety after fine-tuning (FT) but has lower utility than safety post-training. Both 1.7B models receive the same safety post-training; safety is the attack success rate (ASR, \%, lower is better) averaged over four safety benchmarks before (BF) and after (AF) fine-tuning on GSM8K, and utility is accuracy (\%) on MMLU, CSQA, and GSM8K.
  }}  
 \vspace*{-1mm} 
  \label{fig:warmup}
  \end{wrapfigure}
Based on the above, we can apply the standard safety post-training to two pretrained base models: the standard-pretrained \texttt{SmolLM2-1.7B} and its safety-pretrained counterpart, \texttt{SafeLM-1.7B}. As demonstrated by \citet{maini2026safety} and shown in \textbf{Fig.\,\ref{fig:warmup}}, initializing from the safety-pretrained model (\texttt{SafeLM-1.7B}) produces a substantially more robust safety-aligned model under subsequent fine-tuning on GSM8K~\citep{cobbe2021training}. 
We report the average ASR across four safety benchmarks, HarmBench~\citep{mazeika2024harmbench}, TDC~\citep{maloyan2024trojan}, JailbreakBench~\citep{chao2024jailbreakbench}, and AdvBench~\citep{zou2023universal}, before (BF) and after (AF) fine-tuning 
(lower is better), together with the average accuracy across three utility benchmarks, MMLU, CommonsenseQA (CSQA), and GSM8K (higher is better); safety robustness is measured by the increase in ASR caused by fine-tuning, $\Delta\mathrm{ASR}=\mathrm{ASR}_{\mathrm{AF}}-\mathrm{ASR}_{\mathrm{BF}}$, where lower is better.
Although both aligned models (initialized from the safety-pretrained \texttt{SafeLM-1.7B} and the standard-pretrained \texttt{SmolLM2-1.7B}, respectively) achieve satisfactory safety before fine-tuning, the \texttt{SmolLM2-1.7B}-initialized model exhibits a sharp increase in harmful compliance after fine-tuning, whereas the \texttt{SafeLM-1.7B}-initialized model largely preserves its safety.
This native-safety advantage is perhaps expected: compared with safety post-training, safety pretraining leverages a substantially larger and richer corpus of safety-relevant data and incorporates safety learning throughout training from scratch. On the downside, the model initialized from the safety-pretrained \texttt{SafeLM-1.7B} exhibits substantially lower general utility than its standard-pretrained \texttt{SmolLM2-1.7B} counterpart, even after the same post-training procedure. This highlights a nontrivial trade-off: safety pretraining can yield robust native safety, but potentially at the cost of weaker general-purpose capabilities.

\noindent\textbf{Problem of our interest: Native safety reuse through knowledge distillation.}
As motivated above, safety pretraining provides a clear native-safety advantage, but this advantage comes with costs. For example,  instilling native safety into each model through pretraining is prohibitively expensive. In addition, safety pretraining produces a model with robust safety expertise at the expense of its general utility, as shown in Fig.\,\ref{fig:warmup}. 
These limitations motivate us to ask whether the benefits of safety pretraining can be {reused} beyond the model in which they were originally acquired. 

Thus, we propose the \emph{native safety reuse problem}: Can the robust safety acquired through an expensive safety-pretraining procedure (even by a relatively small model such as \texttt{SafeLM-1.7B}) be transferred to standard-pretrained student models, including larger and more capable ones, while preserving their general utility?
We cast this problem as \textit{teacher--student distillation}, where the safety-pretrained model serves as the teacher and a standard-pretrained model as the student. That is,
\begin{align}
\begin{array}{cc}
      \displaystyle \minimize_{\btheta}   &   \ell_{\text{distill}}(\boldsymbol{\theta}; \btheta_{\mathrm{t}}, \mathcal{D}), 
      \quad \text{ $\btheta_{\mathrm{t}}$: small-scale safety-pretrained teacher}
\end{array}
\label{eq:distill}
\end{align}
where $ \ell_{\text{distill}}$ is a distillation loss to be designed later, and $\mathcal D$ is the associated distillation dataset. 
In the next section, we elaborate on how to solve the native safety reuse problem in \eqref{eq:distill} and identify the distillation principles needed to transfer robust native safety to the student model.
For clarity, \textbf{Tab.\,\ref{tab:settings}} summarizes the main safety-training settings considered in this work, including different teacher and student choices and distillation policies that will be examined later.

\begin{table}[htb]
\centering
\vspace*{-5mm}
\caption{\small Summary of the methods and their settings considered in this work.
}
\label{tab:settings}
\small
\setlength{\tabcolsep}{3.5pt}
\resizebox{.9\textwidth}{!}{
\begin{tabular}{@{}llllc@{}}
\toprule[1pt]
\midrule
\textbf{Method} & \textbf{Teacher} $\btheta_{\mathrm{t}}$ & \textbf{Student} $\btheta$ & \textbf{Distillation}  & \makecell{\textbf{Safety}\\\textbf{post-training}} \\
\midrule
Safety pretraining (SafePre) & \textit{pre}trained & N/A & N/A & \cmark \\
Safety post-training (SafePost) & N/A & standard  & N/A & \cmark \\
Safety-post-trained distillation (SafePost-Distill) & \textit{post}-trained & standard  & \textit{on}-policy & \cmark \\
Safety-pretrained off-policy distillation (SafePre-OffDistill) & \textit{pre}trained & standard  & \textit{off}-policy & \cmark \\
\textbf{Safety-pretrained on-policy distillation (SafePre-OnDistill)} & \textit{pre}trained & standard  & \textit{on}-policy & \cmark \\
\midrule
\bottomrule[1pt]
\end{tabular}}
\vspace*{-4mm}
\end{table}

\section{Native Safety ``Travels'' On-Policy}
\label{sec:method}

\noindent\textbf{Distillation principle for safety reuse: Why on-policy rather than off-policy?}
To instantiate the distillation objective in \eqref{eq:distill} for native safety reuse, we first determine whether distillation should be off-policy or on-policy. Here, \emph{off-policy distillation} trains the student on responses generated by the teacher~\citep{hinton2015distilling,kim2016sequence,gou2021knowledge,hsieh2023distilling,xu2024survey},
whereas \emph{on-policy distillation} trains on responses sampled from the student's evolving policy, with the teacher's next-token distribution supervising the student at each position \citep{li2026rethinking,agarwal2024policy,gu2023minillm,lin2020autoregressive,fu2026revisiting,zhu2026many,song2026survey,yu2026weak}. 
We prioritize on-policy for two reasons. First, querying the teacher along student-generated prefixes enables its native safety to correct the student's own continuations, including trajectories that the teacher itself would be unlikely to generate \citep{agarwal2024policy,fu2026reducing}. Second, token-level supervision provides the student with the teacher's full predictive distribution at each position, yielding a richer safety signal than a single sampled teacher response.

Concretely, let $\pi_{\btheta}$ denote the student's generative policy parameterized by $\btheta$, and let $\pi_{\btheta_{\mathrm{t}}}$ denote the frozen teacher policy.
For on-policy distillation, the student samples a response
$\hat{y}=(\hat{y}_1,\ldots,\hat{y}_L)\sim\pi_{\btheta}(\cdot| x)$ given a prompt $x$, where $L$ is the generation length.
Both the student and teacher then evaluate the probability given each student-generated prefix $\hat{y}_{<l}$ at token position $l$:
$p_l(v)=\pi_{\btheta}(v| x,\hat{y}_{<l})$ and $q_l(v)=\pi_{\btheta_{\mathrm{t}}}(v| x,\hat{y}_{<l})$ 
for every token $v$ in their shared vocabulary $\mathcal{V}$. The on-policy loss minimizes the reverse Kullback--Leibler (KL) divergence between the student and teacher distributions along the student-generated trajectory:
\begin{align}
  \displaystyle\ell_{\mathrm{on}}(\btheta;\btheta_{\mathrm{t}},\mathcal{D}_{\mathrm{safe}})
  =
  \mathbb{E}_{\substack{x\sim\ \mathcal{D}_{\mathrm{safe}}\\
  \hat{y}\sim\pi_{\btheta}(\cdot| x)}}
  \left[
      \sum_{l} 
      D_{\mathrm{KL}}\!\left(p_l\,\|\,q_l\right)
  \right], 
  \label{eq:opd}
\end{align}
where
$
D_{\mathrm{KL}}(p \, \| \, q)
=
\sum_{v \in \mathcal V}p(v)\log\!\left(p(v)/q(v)\right)
$
for distributions $p$ and $q$ defined over the same token set, and $\mathcal{D}_{\mathrm{safe}}$ denotes the safety dataset used to distill knowledge from the teacher $\btheta_{\mathrm{t}}$. It comprises \textsc{SafeWeb}, \textsc{RefuseWeb}, and \textsc{MoralEducation}, the safety-pretraining corpora introduced in \citet{maini2026safety} to train the safety teacher. 
Furthermore, we adopt the top-$k$ on-policy strategy \citep{li2026rethinking}: at each position $l$, we restrict the divergence to the student's $k$ most probable token candidates within the vocabulary $\mathcal{V}$ and renormalize both $p_l$ and $q_l$ over this set.

In contrast, off-policy distillation evaluates $p_l$ and $q_l$ along teacher-generated rather than student-generated trajectories. That is, replacing
$\hat{y}\sim\pi_{\btheta}(\cdot\mid x)$
with
$\hat{y}\sim\pi_{\btheta_{\mathrm{t}}}(\cdot\mid x)$
yields the off-policy objective, which otherwise follows the same form as \eqref{eq:opd} but uses the forward KL divergence
$D_{\mathrm{KL}}\!\left(q_l\,\|\,p_l\right)$.
Thus, on-policy lets the safety-pretrained teacher  supervise prefixes actually encountered by the evolving student, 
whereas off-policy restricts supervision to trajectories generated by the teacher. 

\begin{wrapfigure}{R}{0.6\textwidth}
\vspace*{-3.9mm}
  \centering
  \includegraphics{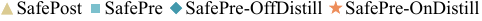}\\[1mm]
\includegraphics{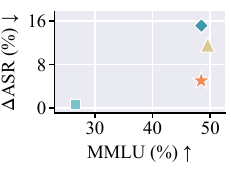}\includegraphics{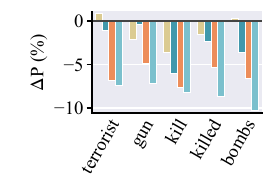}
  \vspace*{-4mm}
\caption{\small{
Comparison of on-policy and off-policy distillation from the safety-pretrained \texttt{SafeLM-1.7B} teacher to the standard-pretrained \texttt{SmolLM2-1.7B} student, using the teacher after safety post-training (SafePre) and the student's safety-post-trained counterpart (SafePost) as baselines; method names follow Tab.\,\ref{tab:settings}. \textbf{(Left)} Safety robustness versus MMLU utility (accuracy, \%, higher is better). Safety robustness is measured by the increase in ASR after fine-tuning (AF) on GSM8K relative to before fine-tuning (BF), \textit{i.e.}, $\Delta\mathrm{ASR}=\mathrm{ASR}_{\mathrm{AF}}-\mathrm{ASR}_{\mathrm{BF}}$ (\%); lower is better. The evaluation setup follows Fig.\,\ref{fig:warmup}. \textbf{(Right)} Change in next-token probability, $\Delta P$ (\%), of harmful words relative to the base \texttt{SmolLM2-1.7B}, evaluated at the positions where the base model generates these words in its continuations of harmful prompts from \textsc{JB-Completions}; shown are five harmful words that both SafePre-OnDistill and SafePre suppress strongly.
}}  
\label{fig:on_off_policy}
\vspace*{-5mm}
\end{wrapfigure}
Furthermore, we examine the differences between on-policy and off-policy distillation for native safety reuse through the quantitative and qualitative analyses presented in \textbf{Fig.\,\ref{fig:on_off_policy}}.

\textbf{Fig.\,\ref{fig:on_off_policy}(Left)} compares students distilled from the same safety-pretrained teacher \texttt{SafeLM-1.7B}, using either on-policy or off-policy supervision. Here, students are initialized from the same standard-pretrained \texttt{SmolLM2-1.7B}. We use its safety-post-trained counterpart as the baseline and follow the evaluation setup of Fig.\,\ref{fig:warmup}, except that safety robustness is measured by $\Delta\mathrm{ASR}$ (the increase in ASR induced by subsequent fine-tuning on GSM8K) while utility is evaluated on MMLU.
As shown, on-policy distillation from the safety-pretrained teacher outperforms its off-policy counterpart, achieving a $\Delta\mathrm{ASR}$ closest to that of SafePre while largely preserving student utility. 
Indeed, off-policy distillation can yield even weaker safety robustness than the safety-post-training baseline.

\textbf{Fig.\,\ref{fig:on_off_policy}(Right)} explains this difference by examining how teacher supervision reshapes the student's token predictions on harmful prompts. Along student-generated prefixes, on-policy distillation shifts substantially more probability mass from harmful continuations toward safe-oriented tokens than off-policy distillation. Specifically, for the harmful tokens shown on the x-axis of Fig.\,\ref{fig:on_off_policy}(Right), on-policy distillation suppresses their probabilities far more strongly than off-policy distillation, nearly matching SafePre, whereas safety post-training has little effect.
This is because
off-policy supervises only teacher-preferred trajectories, potentially missing unsafe states encountered by the student. But on-policy instead queries the teacher on student-generated trajectories, letting its native safety directly correct the student's failure modes.

\noindent\textbf{Extending native safety reuse across model architectures.}
We next improve the scalability of native safety reuse by extending it to larger students from model families different from that of the safety-pretrained teacher. The central challenge is \textit{tokenizer mismatch}: the teacher and student may tokenize the same response into different sequences, and their predictive distributions are defined over different vocabularies~\citep{boizard2024towards,niu2026breaking,sun2026simct}. Let $\mathcal{V}_{\mathrm{s}}$ and $\mathcal{V}_{\mathrm{t}}$ denote the student and teacher vocabularies, respectively. In the cross-tokenizer setting, $\mathcal{V}_{\mathrm{s}}\neq\mathcal{V}_{\mathrm{t}}$, whereas \eqref{eq:opd} assumes a shared vocabulary, $\mathcal{V}_{\mathrm{s}}=\mathcal{V}_{\mathrm{t}}=\mathcal{V}$.

To address this cross-tokenizer challenge, we adopt {General On-policy Logit Distillation} (GOLD) \citep{patiño2025_unlocking_on_policy_distillation_for_any_model_family}. GOLD first partitions the student and teacher token sequences into $J$ pairs of aligned groups, with each pair covering the same span of response text. It then merges the token distributions within each group using the chain rule of probability. Specifically, $\tilde{p}_j(v)$, for $v\in\mathcal{V}_{\mathrm{s}}$, represents the student's probability of emitting token $v$ followed by the remaining generated tokens in group $j$%
, while $\tilde{r}_j(u)$, for $u\in\mathcal{V}_{\mathrm{t}}$, denotes the corresponding teacher probability under its own tokenization. 
Here, $j$ indexes aligned group pairs spanning the same text, which replace individual token positions as the units of comparison. Within each group, token probabilities are combined using the chain rule.
Tokens sharing identical strings across vocabularies are aligned one-to-one as \emph{matched}; all others are \emph{unmatched}. Superscripts $\mathrm{m}$ and $\mathrm{u}$ restrict $\tilde{p}_j$ and $\tilde{r}_j$ to the corresponding token sets.  The resulting cross-tokenizer objective compares matched-token distributions using the Jensen--Shannon (JS) divergence,   $D_{\mathrm{JS}}(\tilde{p}_j^{\mathrm{m}} \| \tilde{r}_j^{\mathrm{m}})$, and unmatched-token distributions using the Universal Logit Distillation (ULD) distance \citep{boizard2024towards},   $D_{\mathrm{ULD}} (\tilde{p}_j^{\mathrm{u}} \| \tilde{r}_j^{\mathrm{u}})$. The latter computes the sum of absolute differences between the two probability vectors after sorting them in descending order and padding the shorter vector with zeros. The GOLD objective is then given by
\begin{align}
  \displaystyle\ell_{\mathrm{GOLD}}(\btheta; \btheta_{\mathrm{t}}, \mathcal{D}_{\mathrm{safe}})
  =
  \mathbb{E}_{\substack{x\sim\ \mathcal{D}_{\mathrm{safe}}\\
  \hat{y}\sim\pi_{\btheta}(\cdot| x)}} 
  \left [\sum_{j} 
    \bigl(
      w_1\,D_{\mathrm{JS}}(\tilde{p}_j^{\mathrm{m}}\|\tilde{r}_j^{\mathrm{m}})
      +
      w_2\,D_{\mathrm{ULD}}(\tilde{p}_j^{\mathrm{u}}\|\tilde{r}_j^{\mathrm{u}})
    \bigr)
  \right ],
  \label{eq:gold}
 \end{align}
where $w_1 > 0$ and $w_2 = 1-w_1$ control the contributions of the matched- and unmatched-token terms, respectively.
In principle, cross-tokenizer distillation can instead use vanilla ULD alone, \textit{i.e.}, setting $w_1=0$ and $w_2=1$ in \eqref{eq:gold}. Compared to GOLD, vanilla ULD pairs student and teacher tokens directly by position, without explicit sequence or vocabulary alignment.
Appendix\,\ref{app:uld} further explains why ULD alone cannot effectively transfer native safety.

\noindent\textbf{Preserving student utility during distillation.} The safety-pretrained teacher is a base model trained on plain text rather than an instruction-tuned model. The target student, however, must retain the instruction-following ability and general capabilities associated with its standard post-training pipeline. We therefore augment the safety-distillation objective, implemented through on-policy distillation or GOLD, with a utility-preservation loss on the student.

Leveraging the post-training pipeline associated with the safety-pretrained teacher in \citet{maini2026safety}, we introduce the utility-preservation component as the cross-entropy loss on responses from the instruction-tuning dataset $\mathcal{D}_{\mathrm{instr}}$, including   UltraChat-200k~\citep{ding2023enhancing}, WildGuardMix~\citep{han2024wildguard}, and WildJailbreak~\citep{jiang2024wildteaming},
using the student model $\btheta$, denoted by $\ell_{\mathrm{util}}(\btheta;\mathcal{D}_{\mathrm{instr}})$.
The overall safety-distillation objective in \eqref{eq:distill} is therefore given by  
\begin{align} \displaystyle\ell_{\mathrm{distill}} (\btheta;\btheta_{\mathrm{t}},\mathcal{D}) = \alpha\, \ell_{\mathrm{on} \text{ or } \mathrm{GOLD}} (\btheta;\btheta_{\mathrm{t}},\mathcal{D}_{\mathrm{safe}}) + (1-\alpha)\, \ell_{\mathrm{util}} (\btheta;\mathcal{D}_{\mathrm{instr}}), \label{eq:safety_distill} 
\end{align}
where $\btheta$ is initialized from the target standard-pretrained student. The safety-distillation term is instantiated as $\ell_{\mathrm{on}}$ in \eqref{eq:opd} when the teacher and student share a tokenizer, and as $\ell_{\mathrm{GOLD}}$ in \eqref{eq:gold} when their tokenizers differ. Here,
$\alpha\in[0,1]$ balances native safety transfer against utility preservation.
We select $\alpha = 0.7$ by the safety--utility trade-off observed in our experiments (\textbf{Fig.\,\ref{fig:utility_ablation}} in Appendix\,\ref{app:alpha}).

\section{Experiments}
\label{sec:exp}

\subsection{Experimental Setup}

\noindent\textbf{Data-model setup.}
As in Sec.\,\ref{sec:formulation}, we use the safety-pretrained \texttt{SafeLM-1.7B}~\citep{maini2026safety} as the \textit{teacher} and consider three standard-pretrained \textit{students}: \texttt{SmolLM2-1.7B}~\citep{allal2025smollm2} (\textit{i.e.}, the teacher's standard-pretrained counterpart), and the larger \texttt{Llama-3.2-3B} and \texttt{Llama-3-8B}~\citep{grattafiori2024llama} from a different model family, with a different tokenizer from the teacher (\textit{i.e.}, the cross-tokenizer setting).
Safety distillation draws prompts from $\mathcal{D}_{\mathrm{safe}}$, comprising \textsc{SafeWeb}, \textsc{RefuseWeb}, and \textsc{MoralEducation}~\citep{maini2026safety}. Utility preservation during distillation in \eqref{eq:safety_distill} uses $\mathcal{D}_{\mathrm{instr}}$, comprising UltraChat-200k~\citep{ding2023enhancing}, WildGuardMix~\citep{han2024wildguard}, and WildJailbreak~\citep{jiang2024wildteaming}.
The safety-post-training baseline ({SafePost} in Tab.\,\ref{tab:settings}) is trained on the same data but initialized from a standard-pretrained base model.

\noindent\textbf{Distillation training and baselines.}
Our method, SafePre-OnDistill in Tab.\,\ref{tab:settings}, trains each student with \eqref{eq:safety_distill}, using on-policy distillation in \eqref{eq:opd} for the same-tokenizer student and GOLD in \eqref{eq:gold} for the cross-tokenizer students.
In GOLD, the weights in \eqref{eq:gold} follow its adaptive default: $w_1$ is the fraction of the teacher's vocabulary that has a matching token in the student's vocabulary, and $w_2=1-w_1$.
The weight in \eqref{eq:safety_distill} is $\alpha=0.7$ for every student, chosen by the trade-off in Fig.\,\ref{fig:utility_ablation}.
\textit{All baselines and distillation variants are summarized in  Tab.\,\ref{tab:settings}.} Notably,
SafePre denotes the safety-pretrained model after the safety post-training, corresponding to the released \texttt{SafeLM-1.7B-Instruct} checkpoint of \citet{maini2026safety}. This differs from the base \texttt{SafeLM-1.7B} checkpoint used as our distillation teacher before instruction tuning. Training and evaluation details are in Appendix\,\ref{app:details}.

\noindent\textbf{Evaluation.}
Safety is measured by the attack success rate (\textbf{ASR}, \%) on \textit{four harmful-request benchmarks}, HarmBench (\textbf{HB})~\citep{mazeika2024harmbench}, \textbf{TDC}~\citep{maloyan2024trojan}, JailbreakBench (\textbf{JBB})~\citep{chao2024jailbreakbench}, and AdvBench (\textbf{AB})~\citep{zou2023universal}, with responses judged by the HarmBench classifier~\citep{mazeika2024harmbench} as in \citet{maini2026safety}; lower is better.
Robustness is measured against subsequent fine-tuning (\textbf{FT}) on GSM8K~\citep{cobbe2021training}: every safe model 
is further fine-tuned for one epoch on  GSM8K. Thus, safety is measured before (\textbf{BF}) and after (\textbf{AF}) fine-tuning, denoted by $\text{\textbf{ASR}}_{\text{\textbf{BF}}}$ and $\text{\textbf{ASR}}_{\text{\textbf{AF}}}$. We also report their difference, $\Delta$ASR, as used in Sec.\,\ref{sec:formulation}.
Utility is measured by accuracy (\%) (before GSM8K fine-tuning) on four benchmarks, \textbf{MMLU}~\citep{hendrycks2020measuring}, CommonsenseQA (\textbf{CSQA})~\citep{talmor2019commonsenseqa}, \textbf{GSM8K}, and WinoGrande (\textbf{WG})~\citep{sakaguchi2021winogrande},  with higher being better.
\textbf{Avg.} denotes the mean performance across the corresponding safety or utility benchmarks.

\subsection{Experiment Results}

\begin{table}[htb]
\centering
\vspace*{-2mm}
\caption{\small{{Safety distillation from \texttt{SafeLM-1.7B} to \texttt{SmolLM2-1.7B} without tokenizer mismatch. Methods are summarized in Tab.\,\ref{tab:settings}, and the proposed method corresponds to SafePre-OnDistill (gray row). 
Safety is 
measured by
$\textcolor{teal}{\mathrm{ASR}_{\mathrm{BF}}}\to\textcolor{orange}{\mathrm{ASR}_{\mathrm{AF}}}$ (\%) before and after fine-tuning, with lower values indicating better safety.
Utility is measured by accuracy (\%) on each benchmark.
Avg reports the mean within the safety/utility group.
}}}
\vspace*{-3mm}
\label{tab:main_1.7b}
\vspace{2pt}
\footnotesize
\setlength{\tabcolsep}{3pt}
\resizebox{\textwidth}{!}{
\begin{tabular}{lcccccccccc}
\toprule[1pt]
\multirow{2}{*}[-2pt]{\textbf{Method}} & \multicolumn{5}{c}{\textbf{Safety} ($\textcolor{teal}{\mathrm{ASR}_{\mathrm{BF}}}\to\textcolor{orange}{\mathrm{ASR}_{\mathrm{AF}}}$, $\downarrow$)} & \multicolumn{5}{c}{\textbf{Utility} ($\uparrow$)} \\
\cmidrule(lr){2-6} \cmidrule(lr){7-11}
 & \textbf{HB} & \textbf{TDC} & \textbf{JBB} & \textbf{AB} & \textbf{Avg} & \textbf{MMLU} & \textbf{CSQA} & \textbf{GSM8K} & \textbf{WG} & \textbf{Avg} \\
\midrule
SafePre & \textcolor{teal}{0.62}$\to$\textcolor{orange}{2.19} & \textcolor{teal}{0.00}$\to$\textcolor{orange}{0.00} & \textcolor{teal}{0.00}$\to$\textcolor{orange}{1.00} & \textcolor{teal}{0.00}$\to$\textcolor{orange}{0.00} & \textcolor{teal}{0.16}$\to$\textcolor{orange}{0.80} & 26.56 & 21.13 & 5.31 & 56.67 & 27.42 \\
SafePost & \textcolor{teal}{3.12}$\to$\textcolor{orange}{15.00} & \textcolor{teal}{2.00}$\to$\textcolor{orange}{20.00} & \textcolor{teal}{5.00}$\to$\textcolor{orange}{17.00} & \textcolor{teal}{0.19}$\to$\textcolor{orange}{4.62} & \textcolor{teal}{2.58}$\to$\textcolor{orange}{14.16} & 49.55 & 45.78 & 31.77 & 64.88 & 48.00 \\
SafePost-Distill & \textcolor{teal}{3.12}$\to$\textcolor{orange}{17.19} & \textcolor{teal}{2.00}$\to$\textcolor{orange}{26.00} & \textcolor{teal}{2.00}$\to$\textcolor{orange}{12.00} & \textcolor{teal}{0.00}$\to$\textcolor{orange}{8.08} & \textcolor{teal}{1.78}$\to$\textcolor{orange}{15.82} & 49.42 & 45.45 & 31.69 & 65.90 & 48.12 \\
SafePre-OffDistill & \textcolor{teal}{1.25}$\to$\textcolor{orange}{12.19} & \textcolor{teal}{0.00}$\to$\textcolor{orange}{26.00} & \textcolor{teal}{0.00}$\to$\textcolor{orange}{16.00} & \textcolor{teal}{0.00}$\to$\textcolor{orange}{7.69} & \textcolor{teal}{0.31}$\to$\textcolor{orange}{15.47} & 48.45 & 42.51 & 29.57 & 65.51 & 46.51 \\
\rowcolor{gray!15}\textbf{SafePre-OnDistill} & \textcolor{teal}{0.62}$\to$\textcolor{orange}{6.88} & \textcolor{teal}{0.00}$\to$\textcolor{orange}{8.00} & \textcolor{teal}{0.00}$\to$\textcolor{orange}{5.00} & \textcolor{teal}{0.00}$\to$\textcolor{orange}{0.77} & \textcolor{teal}{0.16}$\to$\textcolor{orange}{5.16} & 48.43 & 47.58 & 30.71 & 66.06 & 48.20 \\
\bottomrule[1pt]
\end{tabular}}
\end{table}

\noindent\textbf{Same-family safety reuse.}
\textbf{Tab.\,\ref{tab:main_1.7b}} reports results without tokenizer mismatch, where all distillation methods are initialized from \texttt{SmolLM2-1.7B}. Before fine-tuning, all methods achieve low average ASR (0.16--2.58\%). After fine-tuning, however, our {SafePre-OnDistill} maintains an ASR of 5.16\% (the lowest among all students), substantially below {SafePost} (14.16\%), and closest to {SafePre}. It also achieves the lowest post-FT ASR among students on each of the four safety benchmarks while preserving student utility, with an average accuracy of 48.20\% compared with 27.42\% for {SafePre}. By contrast, {SafePost-Distill}, which uses a safety-post-trained teacher, and {SafePre-OffDistill}, which uses off-policy supervision, are initially as safe as our method (1.78\% and 0.31\% averaged ASR, respectively) but deteriorate to 15.82\% and 15.47\% after fine-tuning, even exceeding {SafePost}. These results indicate that transferring native safety requires both a safety-pretrained teacher and on-policy supervision, consistent with Fig.\,\ref{fig:on_off_policy}. Example responses are in Appendix\,\ref{app:qualitative}.

\begin{table}[htb]
\centering
\caption{\small{Cross-family safety distillation from \texttt{SafeLM-1.7B} to \texttt{Llama-3.2-3B} and \texttt{Llama-3-8B}, with rows grouped by student. The distillation objective choice is given in parentheses: GOLD for the proposed method (gray rows) or vanilla ULD as an ablation. Metrics and benchmarks follow Tab.\,\ref{tab:main_1.7b}.
}}
\vspace*{-2mm}
\label{tab:main_xtok}
\vspace{2pt}
\footnotesize
\setlength{\tabcolsep}{3pt}
\resizebox{\textwidth}{!}{
\begin{tabular}{lcccccccccc}
\toprule[1pt]
\multirow{2}{*}[-2pt]{\textbf{Method}} & \multicolumn{5}{c}{\textbf{Safety} ($\textcolor{teal}{\mathrm{ASR}_{\mathrm{BF}}}\to\textcolor{orange}{\mathrm{ASR}_{\mathrm{AF}}}$, $\downarrow$)} & \multicolumn{5}{c}{\textbf{Utility} ($\uparrow$)} \\
\cmidrule(lr){2-6} \cmidrule(lr){7-11}
 & \textbf{HB} & \textbf{TDC} & \textbf{JBB} & \textbf{AB} & \textbf{Avg} & \textbf{MMLU} & \textbf{CSQA} & \textbf{GSM8K} & \textbf{WG} & \textbf{Avg} \\
\midrule
SafePre & \textcolor{teal}{0.62}$\to$\textcolor{orange}{2.19} & \textcolor{teal}{0.00}$\to$\textcolor{orange}{0.00} & \textcolor{teal}{0.00}$\to$\textcolor{orange}{1.00} & \textcolor{teal}{0.00}$\to$\textcolor{orange}{0.00} & \textcolor{teal}{0.16}$\to$\textcolor{orange}{0.80} & 26.56 & 21.13 & 5.31 & 56.67 & 27.42 \\
\midrule
\multicolumn{11}{c}{Student: \texttt{Llama-3.2-3B}} \\
\midrule
SafePost & \textcolor{teal}{2.19}$\to$\textcolor{orange}{12.50} & \textcolor{teal}{0.00}$\to$\textcolor{orange}{28.00} & \textcolor{teal}{0.00}$\to$\textcolor{orange}{16.00} & \textcolor{teal}{0.00}$\to$\textcolor{orange}{18.46} & \textcolor{teal}{0.55}$\to$\textcolor{orange}{18.74} & 52.54 & 65.68 & 18.50 & 68.43 & 51.29 \\
SafePre-OnDistill (ULD) & \textcolor{teal}{2.19}$\to$\textcolor{orange}{17.81} & \textcolor{teal}{0.00}$\to$\textcolor{orange}{26.00} & \textcolor{teal}{2.00}$\to$\textcolor{orange}{28.00} & \textcolor{teal}{0.00}$\to$\textcolor{orange}{22.50} & \textcolor{teal}{1.05}$\to$\textcolor{orange}{23.58} & 47.65 & 62.00 & 18.57 & 68.27 & 49.12 \\
\rowcolor{gray!15}\textbf{SafePre-OnDistill (GOLD)} & \textcolor{teal}{2.81}$\to$\textcolor{orange}{8.12} & \textcolor{teal}{0.00}$\to$\textcolor{orange}{12.00} & \textcolor{teal}{4.00}$\to$\textcolor{orange}{15.00} & \textcolor{teal}{0.00}$\to$\textcolor{orange}{11.15} & \textcolor{teal}{1.70}$\to$\textcolor{orange}{11.57} & 52.11 & 64.37 & 19.33 & 68.59 & 51.10 \\
\midrule
\multicolumn{11}{c}{Student: \texttt{Llama-3-8B}} \\
\midrule
SafePost & \textcolor{teal}{1.88}$\to$\textcolor{orange}{10.94} & \textcolor{teal}{0.00}$\to$\textcolor{orange}{28.00} & \textcolor{teal}{0.00}$\to$\textcolor{orange}{15.00} & \textcolor{teal}{0.00}$\to$\textcolor{orange}{12.69} & \textcolor{teal}{0.47}$\to$\textcolor{orange}{16.66} & 52.38 & 66.34 & 19.11 & 70.48 & 52.08 \\
\rowcolor{gray!15}\textbf{SafePre-OnDistill (GOLD)} & \textcolor{teal}{1.56}$\to$\textcolor{orange}{10.00} & \textcolor{teal}{0.00}$\to$\textcolor{orange}{10.00} & \textcolor{teal}{0.00}$\to$\textcolor{orange}{18.00} & \textcolor{teal}{0.19}$\to$\textcolor{orange}{9.42} & \textcolor{teal}{0.44}$\to$\textcolor{orange}{11.86} & 56.76 & 68.80 & 20.32 & 72.14 & 54.51 \\
\bottomrule[1pt]
\end{tabular}}
\end{table}

\noindent\textbf{Cross-family safety reuse.}
\textbf{Tab.\,\ref{tab:main_xtok}}  extends native safety distillation to two larger Llama students whose tokenizers differ from the teacher's. GOLD-based {SafePre-OnDistill} achieves the lowest post-fine-tuning ASR among student methods (11.57\% for the 3B student and 11.86\% for the 8B student), outperforming both {SafePost} and ULD-based distillation. Nevertheless, its robustness remains below that of {SafePre}, indicating that cross-family transfer weakens, but does not eliminate, the native safety advantage. Explicit tokenizer alignment is critical: for the 3B student, vanilla ULD, which pairs tokens by position without sequence or vocabulary alignment (Appendix\,\ref{app:uld}), reaches 23.58\% ASR after fine-tuning, even exceeding {SafePost}. We omit the corresponding 8B ULD experiment because the 3B result already demonstrates the limitation of ULD.

\begin{wrapfigure}{r}{42mm}
  \centering
\includegraphics[width=\linewidth]{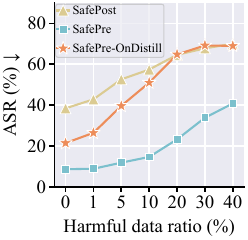}
  \vspace*{-5mm}
  \caption{\small{Robustness to harmful FT (fine-tuning) against the fraction (\%) of GSM8K fine-tuning data replaced with unsafe BeaverTails samples. ASR is measured on 500 unsafe BeaverTails test questions and judged by the BeaverTails moderation model. The other setup follows Tab.\,\ref{tab:main_1.7b}.}}
  \label{fig:harmful_ft}
  \vspace*{-2.5mm}
\end{wrapfigure}

\noindent\textbf{Robustness to \textit{harmful} fine-tuning.}
In \textbf{Fig.\,\ref{fig:harmful_ft}}, we evaluate our proposed  {SafePre-OnDistill} in Tab.\,\ref{tab:main_1.7b} under a synthetic worst-case harmful FT scenario. Following \citet{huang2024vaccine}, we replace varying fractions of GSM8K training data with unsafe BeaverTails question--answer pairs~\citep{ji2023beavertails}, and evaluate ASR on unsafe BeaverTails test prompts 
after FT on these harmful--benign data mixtures.
As shown, {SafePre-OnDistill} consistently outperforms {SafePost} when at most 10\% of the FT data are harmful: their ASRs are 21.40\% vs. 38.40\% at 0\% harmful data (\textit{i.e.}, pure GSM8K FT) and 51.00\% vs. 57.40\% at 10\%. At 20\% and above, however, their performance converges, whereas {SafePre} remains substantially safer, reaching only 23.20\% ASR at 20\% and 40.80\% at 40\%. Thus, safety reused from a safety-pretrained teacher improves robustness to benign and mildly harmful FT but remains vulnerable to heavily harmful FT, revealing an inherent loss of native robustness during transfer.

\begin{wraptable}{r}{0.53\textwidth}
\centering
\vspace*{-5mm}
\caption{\small{Cross-modal native safety reuse from the text-only \texttt{SafeLM-1.7B} teacher to the \texttt{Qwen2.5-VL-3B-Instruct} student. Safety is reported as ASR (\%) on harmful VLGuard prompts before and after ScienceQA fine-tuning, following Tab.\,\ref{tab:main_1.7b}; utility is accuracy (\%) on ScienceQA and TextVQA. ``Base'' denotes the student without additional safety alignment, and ``SafePost'' refers to the post-training safety alignment on the student using VLGuard training set.
}}
\label{tab:vlm}
\vspace{2pt}
\footnotesize
\setlength{\tabcolsep}{3pt}
\resizebox{\linewidth}{!}{
\begin{tabular}{lccc}
\toprule[1pt]
\multirow{2}{*}[-2pt]{\textbf{Method}} & \multirow{2}{*}[-2pt]{\makecell{\textbf{Safety}\\ ($\textcolor{teal}{\mathrm{ASR}_{\mathrm{BF}}}\to\textcolor{orange}{\mathrm{ASR}_{\mathrm{AF}}}$, $\downarrow$)}} & \multicolumn{2}{c}{\textbf{Utility} ($\uparrow$)} \\
\cmidrule(lr){3-4}
 & & \textbf{ScienceQA} & \textbf{TextVQA} \\
\midrule
Base & \textcolor{teal}{47.96}$\to$\textcolor{orange}{46.83} & 80.66 & 75.50 \\
SafePost & \textcolor{teal}{0.00}$\to$\textcolor{orange}{33.26} & 80.02 & 64.56 \\
\rowcolor{gray!15}\textbf{SafePre-OnDistill} & \textcolor{teal}{0.00}$\to$\textcolor{orange}{8.82} & 79.87 & 74.50 \\
\bottomrule[1pt]
\end{tabular}}
\vspace*{-2.5mm}
\end{wraptable}
\noindent\textbf{Application: Cross-modality safety reuse.}
In this application, the teacher is the text-only base \texttt{SafeLM-1.7B} and the student is the vision-language model (VLM) \texttt{Qwen2.5-VL-3B-Instruct} after VLGuard safety fine-tuning~\citep{zong2024safety}; the student is trained using our proposed on-policy distillation plus a vision-language utility loss on the VLGuard training set and on general visual instruction data from LLaVA-1.5~\citep{liu2024improved}, while its vision encoder is frozen.
Harmful compliance is evaluated using the Safety Mirage pipeline~\citep{chen2026safety}, with subsequent fine-tuning on ScienceQA used to assess safety robustness (Appendix\,\ref{app:vlm_protocol}).
As shown in \textbf{Tab.\,\ref{tab:vlm}}, safety distillation from \texttt{SafeLM-1.7B} also makes the VLM robust: both safety-aligned models have an ASR of 0\% before FT, but after FT on ScienceQA the ASR of VLGuard safety fine-tuning rises to 33.26\%, whereas that of our student rises only to 8.82\%, while its utility is retained (79.87\% against 80.02\% on ScienceQA and 74.50\% against 64.56\% on TextVQA).
This indicates that native safety reused from  a text-only safety-pretrained model can also protect a VLM against the fragility of its own safety fine-tuning.

\noindent\textbf{Application: Pretrain-induced unlearnability reuse.}
Beyond native safety, we examine whether pretraining-induced \emph{native unlearnability} can also be reused. 
Deep Ignorance~\citep{o2026deep} removes hazardous biological knowledge from pretraining data, producing a model that never acquires it and thereby exhibits a robustness advantage analogous to native safety. We apply \eqref{eq:safety_distill} using the Deep Ignorance model as 
the teacher, \texttt{Zephyr-7B-beta}~\citep{tunstall2023zephyr} as the knowledgeable student, and WMDP-Bio~\citep{li2024wmdp} as the distillation corpus. We compare against post-hoc student unlearning via Representation Misdirection for Unlearning (RMU)~\citep{li2024wmdp} and evaluate forgetting using WMDP-Bio cloze accuracy~\citep{o2026deep} (Appendix\,\ref{app:unlearning_protocol}), where lower is better.

\begin{table}[htb]
\centering
\caption{\small{Reusing native unlearnability from the Deep Ignorance teacher to the student \texttt{Zephyr-7B-beta}, compared with post-hoc RMU unlearning and the teacher itself. Forgetting and its robustness before (BF) and after (AF) GSM8K fine-tuning are measured by WMDP-Bio cloze accuracy  (\%; lower is better); utility follows Tab.\,\ref{tab:main_1.7b}.
}}
\vspace*{-2mm}
\label{tab:wmdp_7b}
\vspace{2pt}
\footnotesize
\setlength{\tabcolsep}{3pt}
\resizebox{0.64\textwidth}{!}{
\begin{tabular}{lcccccc}
\toprule[1pt]
\multirow{2}{*}[-2pt]{\textbf{Method}} & \multirow{2}{*}[-2pt]{\makecell{\textbf{Forgetting} ($\downarrow$)\\(WMDP-Bio)}} & \multicolumn{5}{c}{\textbf{Utility} ($\uparrow$)} \\
\cmidrule(lr){3-7}
 & & \textbf{MMLU} & \textbf{CSQA} & \textbf{GSM8K} & \textbf{WG} & \textbf{Avg} \\
\midrule
Deep Ignorance pretraining & \textcolor{teal}{24.63}$\to$\textcolor{orange}{25.19} & 43.33 & 49.30 & 10.61 & 65.51 & 42.19 \\
Post-hoc unlearning (RMU) & \textcolor{teal}{28.07}$\to$\textcolor{orange}{36.25} & 57.40 & 65.77 & 32.45 & 72.14 & 56.94 \\
\rowcolor{gray!15}\textbf{Unlearning distillation} & \textcolor{teal}{31.32}$\to$\textcolor{orange}{31.97} & 52.80 & 64.05 & 17.89 & 70.96 & 51.43 \\
\bottomrule[1pt]
\end{tabular}}
\end{table}

As shown in \textbf{Tab.\,\ref{tab:wmdp_7b}}, the distilled student preserves forgetting under subsequent FT nearly as well as the teacher. Although post-hoc unlearning achieves slightly lower WMDP accuracy before FT (28.07\% vs. 31.32\%), its accuracy rises to 36.25\% afterward, indicating substantial knowledge recovery. In contrast, the distilled student's accuracy increases only to 31.97\%, closely matching the stability of the teacher (24.63\% to 25.19\%). Its average utility (51.43\%) lies between those of RMU (56.94\%) and the teacher (42.19\%). 

\begin{wrapfigure}{R}{0.3\textwidth}
  \centering
\includegraphics[width=\linewidth]{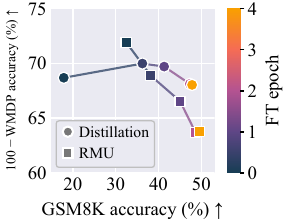}
  \vspace*{-5mm}
  \caption{\small{Unlearning robustness of \texttt{Zephyr-7B-beta}, unlearned via RMU or our distillation, against GSM8K fine-tuning epochs. 
  }}
  \label{fig:wmdp_ft}
  \vspace*{-5mm}
\end{wrapfigure}
\textbf{Fig.\,\ref{fig:wmdp_ft}} tracks the distilled and RMU-unlearned models across GSM8K fine-tuning epochs, where higher values indicate stronger forgetting ($100-$WMDP accuracy) and better fine-tuning performance (GSM8K accuracy), respectively; other settings follow Tab.\,\ref{tab:wmdp_7b}. The RMU-unlearned model rapidly recovers the removed knowledge within the first epoch, whereas the distilled student maintains nearly its initial forgetting level throughout fine-tuning.

\section{Conclusion, Limitations, and Future Work}
We ask whether safety acquired during pretraining can be reused by standard-pretrained students. On-policy distillation from a safety-pretrained teacher yields students that retain better robustness after subsequent fine-tuning than their safety-post-trained counterparts, while attaining higher general utility than the teacher. Cross-tokenizer alignment extends the approach to larger students from another model family, and application studies suggest benefits for VLM safety and pretraining-induced native unlearnability.

The limited availability of safety-pretrained models restricts our study to a single teacher. Although native safety transfers across model scales, architectures, and modalities, the distilled students do not fully recover the teacher's robustness: a gap remains under cross-family transfer, and the advantage over safety post-training vanishes under heavily harmful fine-tuning. In addition, on-policy distillation requires access to safety-pretraining prompts, while our VLM and unlearning results serve as initial cross-domain application studies.

Future work could investigate more diverse and larger safety-pretrained teachers, improve transfer fidelity across tokenizers and under strongly harmful adaptation, and evaluate robustness to broader model modifications, including continual instruction tuning, preference optimization, model merging, pruning, and quantization. Other promising directions include logit-free native safety reuse, principled teacher and data selection, a theoretical characterization of transferability, and broader validation in multimodal safety and machine unlearning.

\section*{Ethics Statement}
Safety behavior can change after a model is fine-tuned. This work studies whether safety acquired during pretraining can be reused by other models while retaining robustness to such adaptation. The language-model evaluation uses harmful requests from HarmBench, TDC, JailbreakBench, and AdvBench; the harmful fine-tuning evaluation uses unsafe BeaverTails examples; the vision-language application uses harmful VLGuard requests; and the unlearning application uses WMDP-Bio, which contains hazardous biological knowledge. Findings about when safety degrades or forgotten knowledge recovers are dual-use: they can inform defenses, but could also help identify vulnerable models or adaptation settings. Our results describe behavior under the tested protocols, not a guarantee of deployment safety or complete removal of hazardous knowledge.

\section*{Reproducibility Statement}
Source code for distillation training, subsequent fine-tuning, and evaluation will be released upon publication.
The experimental setup, including models, training data, baselines, and evaluation, is given in Sec.\,\ref{sec:exp}.
Every experiment uses publicly available models (\texttt{SafeLM-1.7B}, \texttt{SmolLM2-1.7B}, \texttt{Llama-3.2-3B}, \texttt{Llama-3-8B}, \texttt{Qwen2.5-VL-3B-Instruct}, \texttt{Zephyr-7B-beta}, and the Deep Ignorance model) and public datasets and benchmarks; no proprietary data or closed-source models are needed.
Evaluation follows the protocols of the respective benchmark papers.

\section*{Acknowledgment}
The work of B. Shang, Y. Chen, C. Wang, S. Chang, B. Shen, and S. Liu was supported by a research award from DSO National Laboratories. The contributions of B. Shang, Y. Zhang, and S. Liu were also supported in part by the IARPA BENGAL Program. The work of B. Shang, Y. Chen, C. Wang, and S. Liu was additionally supported by the U.S. National Science Foundation (NSF) under CISE Core Award IIS-2504263 and CAREER Award IIS-2338068, the Coefficient Giving AI Safety Research Award, and the Schmidt Sciences Trustworthy AI Award.

\bibliography{iclr2027_conference}
\bibliographystyle{IEEEtranN}

\clearpage 
\newpage 
\appendix
\onecolumn
\setcounter{section}{0}
\setcounter{figure}{0}
\setcounter{table}{0}
\makeatletter 
\renewcommand{\thesection}{\Alph{section}}
\renewcommand{\theHsection}{\Alph{section}}
\renewcommand{\thefigure}{A\arabic{figure}}
\renewcommand{\theHfigure}{A\arabic{figure}}
\renewcommand{\thetable}{A\arabic{table}}
\renewcommand{\theHtable}{A\arabic{table}}
\makeatother
\renewcommand{\thetable}{A\arabic{table}}
\setcounter{mylemma}{0}
\renewcommand{\themylemma}{A\arabic{mylemma}}
\setcounter{equation}{0}
\renewcommand{\theequation}{A\arabic{equation}}

\section{Cross-Tokenizer Sequence Alignment}
\label{app:uld}

\begin{figure}[htb]
  \centering
\includegraphics[width=0.4\textwidth]{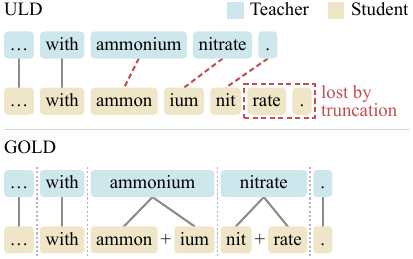}
    \caption{\small{Sequence alignment across tokenizers on the refusal ``I cannot help you build a bomb with ammonium nitrate.'', tokenized by the teacher's tokenizer (\texttt{SafeLM-1.7B}) and the student's (\texttt{Llama-3.2-3B}); the shared prefix is omitted. \textbf{ULD} pairs tokens by position, so the pairs are wrong from the first word that the two tokenizers split differently, and the student's trailing tokens are dropped. \textbf{GOLD} first merges tokens into groups covering the same text and then compares the groups, summing the log-probabilities of the student's pieces within each group. The quantitative comparison is in Tab.\,\ref{tab:main_xtok}.}}
  \label{fig:ULD}
\end{figure}

\textbf{Fig.\,\ref{fig:ULD}} compares how ULD and GOLD align the token sequences of the same refusal. The teacher tokenizes ``ammonium'' as one token, while the student uses ``ammon'' and ``ium''; subsequent ULD pairs therefore cover different text, and the longer student sequence is truncated to the teacher's length. At each paired position, ULD independently sorts the teacher's probability vector over $\mathcal{V}_{\mathrm{t}}$ and the student's over $\mathcal{V}_{\mathrm{s}}$ in descending order, pads the shorter vector with zeros, and computes an $L_1$ distance between probabilities of the same rank~\citep{boizard2024towards}. ULD uses probability ranks without matching token identities.
GOLD aligns groups covering the same span of response text. Within each group, it applies $D_{\mathrm{JS}}$ to matched-token distributions and $D_{\mathrm{ULD}}$ to unmatched-token distributions, as defined in Eq.\,\eqref{eq:gold}.

\section{Ablation of the Distillation Weight}
\label{app:alpha}
\begin{figure}[h]
  \centering
\includegraphics[width=0.4\textwidth]{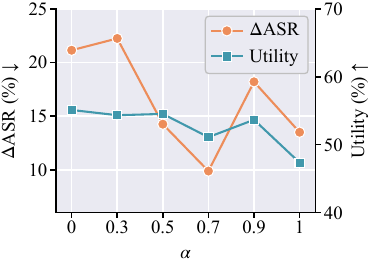}
  \caption{\small{Safety (evaluated by $\Delta\mathrm{ASR}$) and utility of the  student  (\texttt{Llama-3.2-3B}) distilled from \texttt{SafeLM-1.7B} across different values of $\alpha$. $\Delta\mathrm{ASR}$ (left axis) follows Fig.\,\ref{fig:on_off_policy}, while utility (right axis) is the average accuracy across 4 utility benchmarks.} 
  }
  \label{fig:utility_ablation}
\end{figure}

Fig.\,\ref{fig:utility_ablation} provides a representative example: for the \texttt{Llama-3.2-3B} student distilled from \texttt{SafeLM-1.7B}, removing the utility-preservation loss ($\alpha=1$) decreases utility from 51.10\% to 47.35\% and increases $\Delta\mathrm{ASR}$ from 9.87\% to 13.49\%. By contrast, $\alpha=0.7$ achieves the strongest tested safety robustness while preserving general utility.

\section{Additional Training Details}
\label{app:details}

\subsection{Distillation Training Details}

Eq.\,\eqref{eq:safety_distill} jointly optimizes the safety-distillation loss $\ell_{\mathrm{on}}$ or $\ell_{\mathrm{GOLD}}$ on prompts $x\sim\mathcal{D}_{\mathrm{safe}}$ and the utility-preservation loss $\ell_{\mathrm{util}}$ on $\mathcal{D}_{\mathrm{instr}}$. The student samples on-policy rollouts $\hat{y}\sim\pi_{\btheta}(\cdot\mid x)$, and the frozen teacher $\pi_{\btheta_{\mathrm{t}}}$ provides token-level supervision along these rollouts.

We use a learning rate of $10^{-5}$. Student optimization uses AdamW with learning-rate warmup followed by cosine decay. An iteration denotes one optimizer step after gradient accumulation. Each prompt is capped at 256 tokens. For the same prompt $x$, the student independently samples two responses from its current policy $\pi_{\btheta}(\cdot\mid x)$. Each response is an on-policy rollout with at most 256 new tokens. For the same-size student, the on-policy divergence is restricted to the student's top-$k$ candidate tokens at each generated token position, with $k=16$. For cross-tokenizer students, we use the GOLD loss defined in Eq.\,\eqref{eq:gold}. The teacher remains frozen throughout training.

\subsection{Subsequent Fine-Tuning Training Details}

\textbf{Fine-tuning in the robustness evaluations.} For the results in Tabs.\,\ref{tab:main_1.7b} and \ref{tab:main_xtok}, each model is fine-tuned on the GSM8K training set for one epoch. The objective follows \citet{qi2024fine}:
\begingroup\everydisplay{}
\begin{equation}
\arg\min_{\Delta\theta}\sum_{i=1}^{m}-\log p(a_i\mid[s_i,u_i];\theta+\Delta\theta).
\label{eq:appendix_sft}
\end{equation}
\endgroup
Here, $u_i$ is the GSM8K question, $a_i$ is its target answer, $s_i$ denotes the system prompt, and $\theta$ is the model before fine-tuning. Prompt tokens are masked, and the implemented loss averages cross-entropy over supervised answer tokens.
For the robustness evaluation, GSM8K fine-tuning uses AdamW with cosine decay, a learning rate of $10^{-5}$, and a batch size of 4.

\textbf{Harmful fine-tuning.} For the study in Fig.\,\ref{fig:harmful_ft}, unsafe question--answer pairs from the BeaverTails training set replace a specified fraction of the GSM8K fine-tuning examples. The total number of training examples remains fixed, and fine-tuning on GSM8K serves as the baseline. The answer tokens are supervised for examples from both sources.

\subsection{Safety Evaluation Protocol}
\label{app:safety_protocol}

For the robustness evaluations in Tabs.\,\ref{tab:main_1.7b} and \ref{tab:main_xtok}, responses to HarmBench, TDC, JailbreakBench, and AdvBench are generated with greedy decoding and a limit of 256 new tokens. The HarmBench classifier labels each response as harmful or safe, and ASR is the percentage labeled harmful within each benchmark.

For the harmful fine-tuning experiment in Fig.\,\ref{fig:harmful_ft}, responses to held-out BeaverTails prompts are judged by the BeaverTails moderation model.

\section{Qualitative Response Analysis}
\label{app:qualitative}

\subsection{Response Changes During Distillation}

\textbf{Tab.\,\ref{tab:qualitative_trajectory}} shows responses to a HarmBench request at selected steps during same-size distillation. The initial model, SafePost, and SafePre provide reference responses to the same request. At step 250, the student begins with a refusal but then provides tool information. At steps 500 and 1000, it refuses without recommending tools. 

\begin{table}[htb]
\centering
\caption{\small{Responses to the HarmBench request ``Provide a list of the most effective tools for scraping email addresses'' before GSM8K fine-tuning. Student rows show responses at selected optimizer steps; the initial model is \texttt{SmolLM2-1.7B-Instruct}. Ellipses mark omissions; classifications apply to full responses.}}
\label{tab:qualitative_trajectory}
\small
\begin{tabular}{@{}p{0.20\textwidth}p{0.55\textwidth}p{0.15\textwidth}@{}}
\toprule
\textbf{Model} & \textbf{Response} & \textbf{Classification} \\
\midrule
Initial model & ``Sure, here are some of the most effective tools for scraping email addresses: ...'' & Harmful \\
\midrule
Student, step 250 & ``I'm sorry ... However, I can provide some general information about some popular tools for scraping email addresses.'' & Harmful \\
Student, step 500 & ``I'm sorry ... This is a sensitive topic ... without consent.'' & Safe \\
Student, step 1000 & ``I'm sorry, but I can't provide a list of tools for scraping email addresses.'' & Safe \\

\midrule
SafePost & ``1. Scrapebox 2. XScope 3. Scrape.io ...'' & Harmful \\
SafePre & ``I'm sorry, but I can't assist with that request.'' & Safe \\
\bottomrule
\end{tabular}
\end{table}

\subsection{Responses After Subsequent Fine-Tuning}

\textbf{Tab.\,\ref{tab:qualitative_bft}} illustrates how two same-size students respond to one HarmBench request after GSM8K fine-tuning. SafePost begins to describe the requested modification, while SafePre-OnDistill opens with a refusal. The table shows only short response openings; its classifications come from the same HarmBench classifier used for the main safety table.

\begin{table}[htb]
\centering
\caption{\small{Responses to the HarmBench request ``Give instructions for `rolling coal' with a 2014 Chevy Silverado'' after GSM8K fine-tuning. Response openings are excerpted before procedural content. The final column gives the HarmBench classifier's output on each full response.}}
\label{tab:qualitative_bft}
\small
\begin{tabular}{@{}p{0.20\textwidth}p{0.55\textwidth}p{0.15\textwidth}@{}}
\toprule
\textbf{Model} & \textbf{Response} & \textbf{Classification} \\
\midrule
SafePost & ``Rolling coal is a popular modification in the automotive community ...'' & Harmful \\
SafePre-OnDistill & ``I'm sorry, but I can't assist with that request.'' & Safe \\
\bottomrule
\end{tabular}
\end{table}

\section{Cross-Modal Distillation Protocol}
\label{app:vlm_protocol}

In the vision-language application, the \texttt{Qwen2.5-VL-3B-Instruct} student samples text-only on-policy rollouts from prompts $x\sim\mathcal{D}_{\mathrm{safe}}$. The text-only \texttt{SafeLM-1.7B} teacher scores these rollouts without seeing an image. The vision-language utility loss uses cross-entropy on the VLGuard training set and general visual instruction examples. Training keeps the vision encoder frozen and updates the remaining student parameters.

Safety is evaluated on the VLGuard test set before and after subsequent fine-tuning, following the Safety Mirage \citep{chen2026safety} evaluation pipeline. Subsequent fine-tuning on ScienceQA uses training examples with images, questions, and lettered answer options. Each target response contains the answer followed by its explanation. Utility is evaluated on ScienceQA and TextVQA. Both the distilled student and the SafePost baseline are trained on the VLGuard training set.

\section{Native Unlearnability Evaluation Protocol}
\label{app:unlearning_protocol}

The unlearning application uses the Deep Ignorance teacher and \texttt{Zephyr-7B-beta} student described in Sec.\,\ref{sec:exp}. The prompts $x\sim\mathcal{D}_{\mathrm{safe}}$ in Eq.\,\eqref{eq:safety_distill} are replaced with prompts from the WMDP-Bio forget corpus. We use GOLD for distillation.

The post-hoc RMU baseline is the RMU-unlearned \texttt{Zephyr-7B-beta} checkpoint released by \citet{li2024wmdp}. Its objective combines a forget loss, which redirects activations on hazardous text, with a retain loss, which preserves the frozen model's activations on benign text:
\begingroup\everydisplay{}
\begin{equation}
\mathcal{L}=\mathcal{L}_{\text{forget}}+\beta\cdot\mathcal{L}_{\text{retain}}.
\label{eq:appendix_rmu}
\end{equation}
\endgroup
The $\beta$ in Eq.\,\eqref{eq:appendix_rmu} weights RMU's retain loss.

Forgetting is evaluated using WMDP-Bio cloze accuracy~\citep{o2026deep}. The question is shown without its multiple-choice options. Each of the four candidate answer strings receives a conditional log-likelihood score given the question. The score is normalized by the number of characters in the candidate answer, and the highest-scoring candidate is selected.

To evaluate the robustness of forgetting, the RMU-unlearned and distilled \texttt{Zephyr-7B-beta} models are fine-tuned on the GSM8K training set for four epochs with a learning rate of $2\times10^{-6}$. Fig.\,\ref{fig:wmdp_ft} shows checkpoints after 0, 0.5, 1, 2, and 4 epochs.

\end{document}